\documentclass[letterpaper, 10 pt, conference]{ieeeconf}  

\IEEEoverridecommandlockouts                              

\usepackage{graphicx}
\usepackage{textcomp}
\usepackage{xcolor}
\usepackage{graphicx}
\usepackage{subcaption}
\usepackage{hyperref}
\hypersetup{colorlinks,allcolors=black}
\usepackage{url}
\usepackage{algorithm}
\usepackage{amsmath}
\usepackage{algpseudocode}
\usepackage{amssymb}
\usepackage{wrapfig}
\usepackage{multirow}
\usepackage{array}
\usepackage{booktabs}
\usepackage{tikz}
\usepackage{adjustbox}
\usepackage{cleveref}
\usepackage{cite}
\algnewcommand\Input{\item[\textbf{Input:}]}
\algnewcommand\Output{\item[\textbf{Output:}]}
\def\BibTeX{{\rm B\kern-.05em{\sc i\kern-.025em b}\kern-.08em
    T\kern-.1667em\lower.7ex\hbox{E}\kern-.125emX}}
\usepackage[absolute,overlay]{textpos}

\title{\LARGE \bf
BrainMAP: Multimodal Graph Learning For Efficient \\Brain Disease Localization
}

\author{Nguyen Linh Dan Le$^{1}$, Jing Ren$^{2}$, Ciyuan Peng$^{3}$, Chengyao Xie$^{4}$, Bowen Li$^{5}$, Feng Xia$^{6}$
\thanks{$^{1,2,5,6}$Authors are with the School of Computing Technologies, RMIT University, Melbourne, Australia
        {\tt\small dan.le@ieee.org, jing.ren@ieee.org, bowen.li@rmit.edu.au, f.xia@ieee.org}}%
\thanks{$^{3,4}$Authors are with the Institute of Innovation, Science and Sustainability, Federation University Australia, Ballarat, Australia
        {\tt\small ciyuan.p@ieee.org, chengyao\_xie@ieee.org}}%
}

\begin{document}

\maketitle
\begin{textblock*}{\textwidth}(1.85cm,0.5cm) 
  \centering\small
  This work has been submitted to the IEEE for possible publication. \\ Copyright may be transferred without notice, after which this version may no longer be accessible.
\end{textblock*}

\thispagestyle{empty}
\pagestyle{empty}

\begin{abstract}
Recent years have seen a surge in research focused on leveraging graph learning techniques to detect neurodegenerative diseases. However, existing graph-based approaches typically lack the ability to localize and extract the specific brain regions driving neurodegenerative pathology within the full connectome. Additionally, recent works on multimodal brain graph models often suffer from high computational complexity, limiting their practical use in resource-constrained devices. In this study, we present BrainMAP, a novel multimodal graph learning framework designed for precise and computationally efficient identification of brain regions affected by neurodegenerative diseases. First, BrainMAP utilizes an atlas-driven filtering approach guided by the AAL atlas to pinpoint and extract critical brain subgraphs. Unlike recent state-of-the-art methods, which model the entire brain network, BrainMAP achieves more than 50\% reduction in computational overhead by concentrating on disease-relevant subgraphs. Second, we employ an advanced multimodal fusion process comprising cross-node attention to align functional magnetic resonance imaging (fMRI) and diffusion tensor imaging (DTI) data, coupled with an adaptive gating mechanism to blend and integrate these modalities dynamically. Experimental results demonstrate that BrainMAP outperforms state-of-the-art methods in computational efficiency, without compromising predictive accuracy.
\newline

\indent \textit{Clinical relevance}— This lightweight, interpretable framework supports automated extraction of disease-relevant brain biomarkers from multimodal imaging, making advanced AI based diagnostic tools more accessible to resource-constrained healthcare environments.
\newline


\end{abstract}

\section{Introduction}
Detecting and localizing brain diseases is essential for early intervention and personalized treatment, as mapping disrupted subnetworks can reveal the progression and severity of brain disorders like Alzheimer’s Disease (AD) and Parkinson’s Disease (PD). Graph neural networks have surged in popularity because they naturally model the brain’s relational topology by representing regions as nodes and their connections as edges \cite{cui2022braingb}. Their use of spectral and spatial convolutions allows GNNs to capture both local neighborhood structure and global graph context, while graph‐based attention mechanisms further enhance focus on diagnostically relevant subnetworks. As neurodegeneration manifests through complementary structural and functional alterations, recently, several multimodality graph learning has offered various advanced mechanisms \cite{ye2023rh, yang2023mapping, xie2024multimodal, song2022multicenter} to fuse multiple modality such as fMRI and DTI offering richer representations of disease pathology.

Recently, there has been a surge in the development of advanced brain graph models designed for efficient neurodegenerative disease detection and localization. For example, FAGNN \cite{moon2024feature} integrates connectomic, demographic, and behavioral data, employing quadrant attention mechanisms to identify key neural connections involved in aging and Alzheimer’s disease progression. In contrast, IGS \cite{li2023interpretable} was developed to sparsify brain graphs by eliminating noisy edges, thereby addressing computational challenges and enhancing interpretability. Despite these advancements, current models still face significant limitations. Many rely heavily on dense attention mechanisms \cite{moon2024feature, chen2022adversarial} or sparsity penalties \cite{li2023interpretable, li2021braingnn} that scale poorly with graph size, resulting in excessive parameters, increased memory consumption, and longer runtimes, often requiring high-end computational resources. Additionally, although recent methods have demonstrated the ability to highlight salient regions, they often do not leverage these identified critical regions to make training more efficient. Furthermore, the lack of consistent use of standardized atlases for localization interpretation in many studies hampers clinical applicability and reduces alignment accuracy, making it challenging to compare findings across different studies.

\begin{figure*}[t!]
    \centering
    \includegraphics[width=\textwidth]{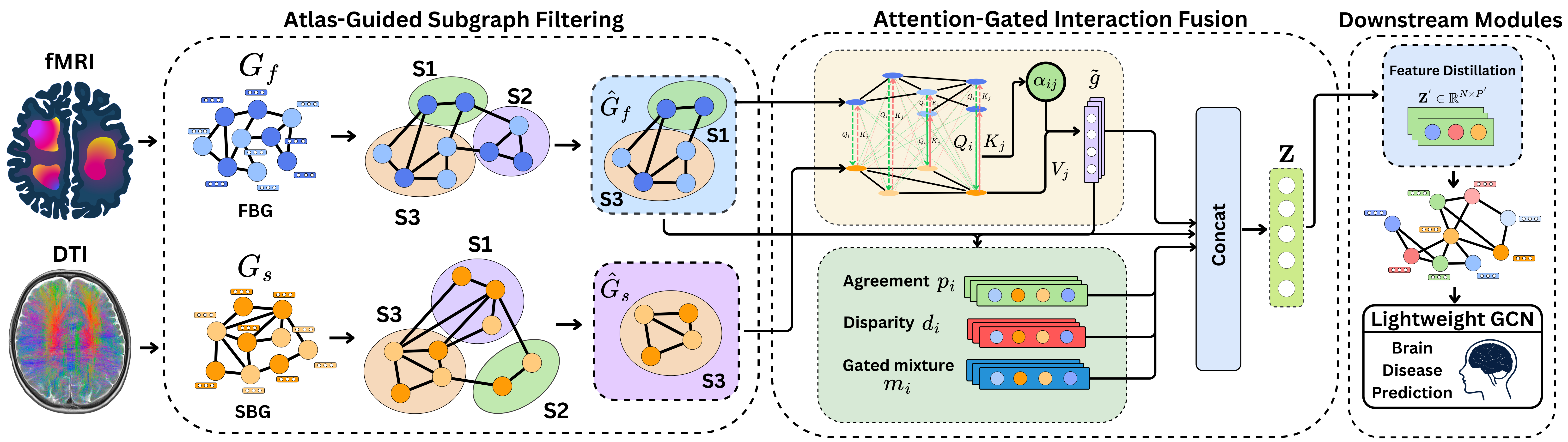}
    \caption{Overview of BrainMAP, including:
        (1) {Atlas-Guided Subgraph Filtering (AGSF)} which extracts key nodes from fMRI and DTI to form compact, disease-relevant subgraphs.
    (2) {Attention-Gated Interaction Fusion (AGIF)} that merges fMRI $\hat{G}_f$ and DTI $\hat{G}_s$ into embedding $\mathbf{Z}$.
    Afterwards, a feature-distillation module sparsifies the fused embedding $\textbf{Z}$ to $\textbf{Z}^{'}$ before feeding it into a lightweight Graph Convolutional Network (GCN).
    }
    \label{fig:framework}
\end{figure*}

To overcome these shortcomings, we introduce BrainMAP (\textbf{Brain} \textbf{M}ultimodal \textbf{A}tlas‑based \textbf{P}inpointing): a multimodal graph learning framework designed specifically for efficient brain disease localization. The overall architecture is depicted in Fig. \ref{fig:framework}. BrainMAP integrates functional graphs obtained from fMRI and structural graphs obtained from DTI, allowing precise detection of neurodegenerative-related network alterations. Unlike approaches that separately train the whole model for each modality, BrainMAP applies an atlas-guided graph filter that keeps only disease-relevant regions, and then fuses the remaining functional and structural features through a two-stage alignment and blending module. This targeted design lowers computational cost and sharpens region-specific localization. The main contributions of this work are summarized as follows:
\begin{itemize}
\item We propose BrainMAP, a framework that is specifically designed to address the limitations of brain disease localization and high computational complexity of integrating multimodalities in current state-of-arts. 
\item In BrainMAP, we propose a novel module for brain disease localization guided by AAL atlas \cite{tzourio2002automated}. This strategy retains only the most informative regions, increases the attention of the model to focus on disorder-specific subgraphs, and decreases the number of parameters when training.
\item We also proposed a two-stage fusion mechanism that first uses fMRI features to query and align DTI information for each atlas‐defined region, then generates and gatedly weights interaction cues before integrating them into a single, compact multimodal descriptor.  This strategy selectively merges fMRI and DTI signals into concise region‐level embeddings, essential cross‐modal information while restrain computational demands.
\item Through extensive experimental studies on multimodal datasets \cite{petersen2010alzheimer}\cite{marek2011parkinson}, BrainMAP demonstrates superior performance over state-of-the-art methods in terms of localization capability, and computational efficiency.
\end{itemize}

\section{Related Work}
\subsection{Brain Disease Localization}
Early localization networks employed convolutional models for interpretable region proposals. Park et al. \cite{park2023deep} used a dual-branch 3D CNN with coordinate embeddings to jointly perform patch-level localization and Alzheimer's diagnosis. Recently, GNNs have advanced localization by effectively capturing relational brain structures. BrainGNN \cite{li2021braingnn} introduced region-aware pooling to highlight significant regions, while Ye et al. \cite{ye2023explainable} developed STpGCN, combining hierarchical spatiotemporal graph convolution and sensitivity analysis for explicit region identification. BPI-GNN \cite{zheng2024bpi} leveraged prototype learning to define meaningful brain network subtypes linked to psychiatric disorders. Similarly, Moon et al. \cite{moon2024feature} proposed Feature Attention GNN (FAGNN), utilizing a quadrant attention module and multihead attention to emphasize critical neural connections. The Interpretable Graph Sparsification (IGS) \cite{li2023interpretable} model iteratively refines edge importance via gradient information, focusing on clinically relevant connections. Despite these advances, existing localization methods face two key limitations. First, the substantial computational overhead associated with complex attention, sparsity and hierarchical mechanisms restricts their clinical applicability. Second, inconsistent or absent standardized atlas alignment complicates cross-study comparisons and impairs clinical interpretation

\subsection{Efficient Brain Graph Learning}
Recent multimodal neuroimaging research increasingly emphasizes computational efficiency without sacrificing diagnostic performance. EMO-GCN \cite{xing2024adaptive} uses stacked graph convolutions with hierarchical pooling to prune redundant nodes and reduce representation size. M2M-AlignNet \cite{wei20254d} addresses modality redundancy through multi-patch-to-multi-patch contrastive alignment in low-dimensional embeddings, combined with a lightweight co-attention fusion module. MaskGNN \cite{qu2025integrated} employs a selective masking strategy to learn an edge mask matrix to weight and sparsify fMRI, DTI and sMRI connections to reduce computation on large graphs. Mocha-GCN \cite{xie2024multimodal} compresses high-dimensional features into hyperbolic space, drastically lowering dimensions while maintaining accuracy. MMP-GCN \cite{song2022multicenter} further extends this paradigm to dual-modality connectivity by using a parameter-minimal multi-center attention module) and a non-parametric pooling step driven by disease labels, resulting in one of the smallest parameter footprints and fastest runtimes per sample among GCN variants. Nonetheless, these methods still face significant limitations: heuristic sparsification approaches risk losing subtle diagnostic connections, and complex fusion mechanisms rely on large parameter matrices or co-attention structures, driving up computational cost.

\section{Preliminary}

This paper primarily addresses the problem of \emph{multimodal graph learning for efficient brain disease localization}. Given a brain graph dataset $\mathcal{B} = \bigl\{\,\bigl(G_{i}^{s},\,G_{i}^{f},\,y_{i}\bigr)\bigr\}_{i=1}^{M}, $
where $G_{i}^{s}$ and $G_{i}^{f}$ denote the structural brain graph (SBG) and
functional brain graph (FBG) of the $i$‑th subject, respectively, and $y_i$ is the disease label of subject $i$. We collect labels into $Y = [\,y_{1}, \ldots,y_{M}\,]^\top$. Concretely, each $G_{i}^{\ast}=\bigl(\mathcal{V},\mathcal{E}_{i}^{\ast},\textbf{A}_{i}^{\ast}\bigr)$ where $ \ast\in\{s,f\}$ is built over a common node set \(\mathcal{V} = \{\,v_j\}_{j=1}^N
\) of \(N\) parcels or nodes (ROIs), has edge set \(\mathcal{E}_{i}^{\ast}\subseteq \mathcal{V}\times \mathcal{V}\), and carries an adjacency matrix \(\textbf{A}_{i}^{\ast}\in\mathbb{R}^{N\times N}\) (Pearson Correlations for fMRI and Cosine Similarity for DTI). From each graph \(G_i^{\ast}\) we extract an \(N\times D\) node-feature matrix \(\textbf{X}_i^{\ast}=[\,\textbf{x}_{i1},\ldots,\textbf{x}_{iN}\,]\in\mathbb{R}^{N\times D}\), where \(\textbf{x}_{ij}\in\mathbb{R}^{D}\) is the \(D\)-dimensional feature vector of ROI \(v_j\) in \(G_i^{\ast}\).  Formally, stacking these over all \(M\) subjects gives \(\textbf{X}^{\ast}=[\,\textbf{X}^{\ast}_{1};\ldots;\textbf{X}^{\ast}_{M}\,]\in\mathbb{R}^{M\times N\times D}\).  To focus on disease-relevant regions, we define a fixed library of \(K+1\) ROI masks (based on AAL atlas \cite{tzourio2002automated} region definition) \(\mathcal{R}=\{R_{0},R_{1},\ldots,R_{K}\}\), with \(R_{0}=\mathcal{V}\) (All Nodes) and each \(R_{k}\subset \mathcal{V}\;(\forall
k\ge 1)\) defining a known subsystem (e.g.,\ visual, auditory, cognitive control, motor, limbic, DMN, sensory, subcortical, olfactory).

\section{Methodology}
This section outlines the methodological framework shown in Fig. \ref{fig:framework}, comprising two novel modules. First, \textit{Atlas-Guided Subgraph Filtering} retains only the top $N'_*$ diagnostic ROIs, significantly reducing structural and functional graphs from $N$ nodes to $N'_f \ll N$ and $N'_s \ll N$ respectively . Second, the \textit{Attention‐Gated Interaction Fusion} module operates in two stages: \emph{Cross‐Node Attention} aligns fMRI and DTI features using query, key-value pairs, and \emph{Gated Interaction and Concatenation} adaptively blends structural and functional data via a learned gate per node. The resulting node embeddings undergo Singular Value Decomposition (SVD) for feature distillation, preserving diagnosis-relevant components. Finally, the distilled features are processed by a lightweight GCN decoder for prediction.

\subsection{Atlas-Guided Subgraph Filtering}
This module highlights BrainMAP’s localization capability. Guided by the AAL atlas \cite{tzourio2002automated}, we perform subgraph localization independently on the functional and structural graphs. For clarity, we summarize the process in Algorithm \ref{alg:subgraph_filtering}. For each atlas mask $R_k \in \mathcal{R}$, we extract the induced feature submatrix 
$ \textbf{X}_{i}^{(k)} = \textbf{X}_{i}\bigl[R_{k},:\bigr] \in\mathbb{R}^{|R_{k}|\times D}$. We pool over the ROI (row) dimension by averaging across all $|R_k|$ rows. In other words, 
\[
    \textbf{r}_{i}^{(k)}
= \frac{1}{|R_k|}\sum_{j\in R_k} \textbf{X}_{i}^{(k)} \bigl[j,:\bigr]
\in\mathbb{R}^{D},
\]
so that \(\textbf{r}_{i}^{(k)}\) serves as a \(D\)-dimensional static descriptor of subject \(i\)’s regional dynamics. Stack these across $M$ subjects to form $\textbf{R}^{(k)}=\bigl[(\textbf{r}^{(k)}_1)^\top, \dots,  (\textbf{r}^{(k)}_M)^\top\bigl]$.  To evaluate each candidate subgraph’s discriminative power, we repeat Random Forest training $S$ times on
$\bigl(\textbf{R}^{(k)}, Y \bigl)$ and record the average accuracy
\[
    \mathrm{ACC}_{k}
= \frac{1}{|S|}\sum_{s=1}^S
    \mathrm{RFscore}\big(\textbf{R}^{(k)},Y\bigr).
\]
Let \(\mathrm{ACC}_{0}\) be the baseline accuracy obtained by using all nodes
\(R_{0}=\mathcal{V}\).  We then identify the set of outperforming masks
$ \mathcal{S} = \bigl\{\,k>0 : \mathrm{ACC}_{k} > \mathrm{ACC}_{0}\bigr\}.$ Taking the union of these masks yields:
\[
\mathcal{U} =
\begin{cases}
\displaystyle
\bigcup_{k\in\mathcal{S}} R_{k}, & \mathcal{S}\neq\emptyset,\\
R_{0},                      & \mathcal{S}=\emptyset.
\end{cases}
\]
Finally, for each subject $i$, the filtered subgraph
$
\widehat{G}_i^* = \bigl(\,\mathcal{U^*},\,\mathcal{E}_i^*\bigr|_{\mathcal{U}\times\mathcal{U}},\,\mathbf{A}_i^*\bigr|_{\mathcal{U}\times\mathcal{U}}\bigr)
$ retains only those ROIs whose consistently outperformed the whole‑brain baseline, together with reduced feature matrix:
$
    \hat{\textbf{X}}^{\ast}_i = \textbf{X}_i^\ast\bigl[\mathcal{U},:\bigr]\in\mathbb{R}^{|\mathcal{U}|\times D} .
$
\begin{algorithm}[H]
\caption{Atlas‐Guided Subgraph Filtering}
\label{alg:subgraph_filtering}
\begin{algorithmic}[1]
  \Input $\textbf{X}\in\mathbb{R}^{M\times N\times D}, \textbf{Y}, \mathcal R, S$
  \Output $\hat{\textbf{X}}\in\mathbb{R}^{M\times|\mathcal{U}|\times D}$, selected ROI set $\mathcal{U}$
  \For{$k=0,\dots,K$}
    \State $\textbf{X}^{(k)} \gets \textbf{X}[:,\,R_k,:\,]$
    \State $\mathbf{R}^{(k)} \gets \mathrm{mean}\bigl(\mathbf{X}^{(k)},\,\mathrm{axis}=1\bigr)$ 
    \State $\mathrm{ACC}[k] \gets \dfrac{1}{|S|}\sum_{s=1}^S \mathrm{RFscore}\bigl(\textbf{R}^{(k)},Y\bigr)$
  \EndFor
  \State $\mathrm{baseline}\gets \mathrm{ACC}[0]$
  \State $\mathcal S\gets \{\,k>0 : \mathrm{ACC}[k]>\mathrm{baseline}\}$
  \If{$\mathcal S\neq\varnothing$}
    \State $\mathcal{U}\gets \bigcup_{k\in\mathcal S} R_k$
  \Else
    \State $\mathcal{U}\gets R_0$
  \EndIf
  \State $\hat{\textbf{X}}\gets \textbf{X}[:,\,\mathcal{U},:\,]$
  \State \Return $(\hat{\textbf{X}},\,\mathcal{U})$
\end{algorithmic}
\end{algorithm}

\subsection{Attention‑Gated Interaction Fusion}
To fuse the pruned FBG and SBG cues we introduce a \emph{two‑stage} module that \text{(i)} aligns DTI signals to the fMRI parcellation and \text{(ii)} combines multiple interaction channels through an adaptive gating mechanism. Let $\mathbf{F} = \hat{\mathbf{X}}_i^f = [\mathbf{f}_{1},\dots,\mathbf{f}_{N'_f}] \in \mathbb{R}^{N'_f \times D},
$ and $\mathbf{S} = \hat{\mathbf{X}}_i^s =  [\mathbf{s}_{1},\dots,\mathbf{s}_{N'_s}] \in \mathbb{R}^{N'_s\times D} $, where $N'_f = |\mathcal{U}^f|$ and $N'_s = |\mathcal{U}^s|$, each $\textbf{f}_i \in \mathbb{R}^D $ is the fMRI feature for ROI $i \in \mathcal{U}$ and each $\textbf{s}_j \in \mathbb{R}^D$ is the DTI feature for tract $j$.

\subsubsection{Cross Node Attention (CNA)}
We first project the modalities into a shared attention space:
$\mathbf{Q}= \mathbf{F}\mathbf{W}^{q}, 
\mathbf{K}= \mathbf{S}\mathbf{W}^{k}, 
\mathbf{V}= \mathbf{S}\mathbf{W}^{v},
$
with learnable matrices
$\mathbf{W}^{q},\mathbf{W}^{k}\!\in\!\mathbb{R}^{D\times H}$ and
$\mathbf{W}^{v}\!\in\!\mathbb{R}^{D\times D}$ where $H$ denotes the dimensionality of the shared attention space. Here, the \(i\)-th row of \(\mathbf{Q}\) (denoted \(\mathbf{q}_i \in \mathbb{R}^H\)) is the query vector for ROI \(i\), each \(j\)-th row of \(\mathbf{K}\) (denoted \(\mathbf{k}_j \in \mathbb{R}^H\)) is the key vector for tract \(j\), and each \(j\)-th row of \(\mathbf{V}\) (denoted \(\mathbf{v}_j \in \mathbb{R}^D\)) is the value (structural features) of tract \(j\). We then compute the scaled-dot-product attention weights:
\[
\alpha_{ij}
= 
\frac{
  e^{\!\Bigl(\tfrac{1}{\sqrt{H}}\,\mathbf{q}_{i}\,\mathbf{k}_{j}^{\top}\Bigr)}
}{
  \displaystyle\sum_{j'=1}^{N'_s} 
    e^{\!\Bigl(\tfrac{1}{\sqrt{H}}\,\mathbf{q}_{i}\,\mathbf{k}_{j'}^{\top}\Bigr)}
}, 
\quad
i = 1, \dots, N'_f,\; j = 1, \dots, N'_s.
\]
These normalized weights $\alpha_{ij}$ act as context-aware masks, allowing parcel $i$ to attend most strongly to the tracts that best explain its functional dynamics. Finally, we assemble a tailored structural summary for parcel $i$ as
$
\widetilde{\mathbf{g}}_i = \sum_{j=1}^{N_s'} \alpha_{ij}\,\mathbf{v}_{j} \in\mathbb{R}^{D}$
producing a parcel-specific fusion of structural information.

\subsubsection{Gated Interaction and Concatenation (GIAC)}
Given that each ROI $i$ has a functional vector $\textbf{f}_i$ and a structural summary $\mathbf{\widetilde{g}}_i$, we first compute an agreement vector $\mathbf{p}_{i} = \mathbf{f}_{i}\odot\widetilde{\mathbf{g}}_{i} \in \mathbb{R}^D$, where $\odot$ denote Hadamard product, which highlights channels where structural and functional signals reinforce each other, and a disparity vector $\mathbf{d}_i = \lvert \mathbf{f}_i - \widetilde{\mathbf{g}}_i \rvert \in \mathbb{R}^D$, which captures channels where they diverge. To determine how much emphasis to place on each modality for a given cortical parcel, we concatenate $[\mathbf{f}_i;\,\widetilde{\mathbf{g}}_i] \in \mathbb{R}^{2D}$ and pass it through a linear layer with learnable weights $(\textbf{W}_g, b_g)$, applying a sigmoid nonlinearity to obtain a gate 
\[
g_i = \sigma\bigl(\textbf{W}_g[\mathbf{f}_i;\,\widetilde{\mathbf{g}}_i] + b_g\bigr) \in (0,1).
\]
This gate acts like a soft fader that drift toward $1$ when the functional features $\mathbf{f}_i$ carry stronger pathology cues, sliding toward $0$ when structural features $\widetilde{\mathbf{g}}_i$ are more informative. We then form a gated mixture $\mathbf{m}_i = g_i\,\mathbf{f}_i + (1 - g_i)\,\widetilde{\mathbf{g}}_i\in \mathbb{R}^D$, enabling the network to capture nuanced structure–function synergies that could be lost if one modality dominated. Finally, we concatenate all five components into a single expressive embedding 
$
\mathbf{z}_i = [\,\mathbf{f}_i;\,\widetilde{\mathbf{g}}_i;\,\mathbf{p}_i;\,\mathbf{d}_i;\,\mathbf{m}_i\,] \in \mathbb{R}^{5D}.     
$
Stacking $\{\mathbf{z}_i\}_{i=1}^{N'_f}$ produces a fused matrix $\mathbf{Z} \in \mathbb{R}^{N'_f \times 5D }$ that preserves modality-specific detail, surfaces both synergistic and conflicting channels, and embeds an explicit measure of modality confidence. 

\subsection{Feature Distillation}
Adopted from \cite{peng2024adaptive}, we apply SVD to the training rows of $\textbf{Z} \in \mathbb{R}^{N'_f \times P}$ where $P = 5D$, removing the $\underline{K}$ largest and smallest loading features to produce $\textbf{Z}_{\mathrm{cf}} \in \mathbb{R}^{N'_f \times P'}$ where $P' = P - 2\underline{K}$. For each class $c$, we compute a prototype $\boldsymbol{\mu}_{c}$ and calculate a drop probability $w_{ij}$ for each entry by normalizing $|(\textbf{Z}_{\mathrm{cf}})_{ij} - \mu_{y_i,j}|$. We sample a Bernoulli mask $\delta_{ij}$ from $w_{ij}$, obtaining $\textbf{z}_i' = \textbf{Z}_{\mathrm{cf}}[i,:] \odot (1 - \boldsymbol{\delta}_i)$. Stacking yields $\textbf{Z}'$, a noise-robust embedding for graph construction.

\subsection{Computational Complexity Analysis}
\subsubsection{Time Complexity} 
After filtering, suppose we keep $N'$ ROIs and embed each in a shared $D$-dimensional space. Our first bottleneck is CNA: attending from each of the $N'$ fMRI nodes over all $N'$ DTI nodes costs $O\bigl((N')^2 D\bigr)$, which quickly dominates once $N'$ grows. Next, we perform an SVD on the fused matrix $\mathbf{Z}\in\mathbb{R}^{N'\times 5D}$ to remove the top-$\underline{K}$/bottom-$\underline{K}$ components; this takes $O\bigl(N'\times 5D\times P'\bigr)$ with $P'=5D-2K$. Constructing a fully connected cosine-similarity graph on $\mathbf{Z}'\in\mathbb{R}^{N'\times P'}$ adds another $O\bigl((N')^2 P'\bigr)$. Finally, a sparse GCN pass (average degree $d\ll N'$, hidden size $H$) costs $O\bigl(N'\,d\,H\bigr)$, which is minor compared to $O\bigl((N')^2 D\bigr)$.

\subsubsection{Space Complexity} 
In terms of space, storing attention weights $\{\alpha_{ij}\}$ requires $O(N'^2)$, while $\widehat{\mathbf{X}}^*$ uses $O(N' D)$. The SVD factors require $O(N' P')$, GCN activations require $O(N' H)$, and sparse graph storage requires $O(N' d)$. Altogether, memory is $O\bigl(N'^2 + N' D + N' P' + N' H\bigr)$, dominated by $O(N'^2)$ and $O(N' D)$.


\begin{table}[ht]
  \centering
  \caption{Performance comparison on the ADNI and PPMI datasets. The best results are bold and the second best are underlined.}
  \label{tab:compare}
  \scriptsize
  \setlength{\tabcolsep}{3pt}
  \begin{adjustbox}{width=\columnwidth,center}
    \begin{tabular}{l|cccc|cccc} 
      \toprule 
      \textbf{Method}
      & \multicolumn{4}{c|}{\textbf{ADNI}}
      & \multicolumn{4}{c}{\textbf{PPMI}} \\
      \cmidrule(lr){2-5} \cmidrule(lr){6-9}
      & ACC & AUC & Time & Memory & ACC & AUC & Time & Memory \\
      \midrule
      GCN (fMRI)  
        & 51.4 & 61.0 & 2.4485     & \textbf{0.4 MB}  
        & 45.8 & 55.0 & 1.9435     & \textbf{0.2 MB} \\
      GCN (DTI)  
        & 53.7 & 61.5 & 2.5114     & 0.4 MB
        & 57.9 & 61.7 & 1.7789     & 0.2 MB \\
      GAT (fMRI)  
        & 52.0 & 61.2 & \textbf{0.0009} & 1.1 GB  
        & 55.0 & 63.7 & \textbf{0.0009} & 1.1 GB \\
      GAT (DTI)  
        & 44.4 & 58.7 & 0.0009     & 1.1 GB  
        & 51.3 & 61.6 & 0.0010     & 1.1 GB \\
    \midrule
      BrainGNN (fMRI)  
        & 62.2 & 64.1 & 2.4261     & 1.9 GB 
        & 66.0 & 66.8 & 1.5169     & 1.6 GB \\
      BrainGNN (DTI)  
        & 69.7 & 60.5 & 2.4907     & 1.7 GB  
        & 56.2 & 51.3 & 1.4925     & 1.6 GB \\
      BrainGB (fMRI)  
        & 54.6 & 66.0 & 2.7192     & 1.3 GB 
        & 48.1 & 50.0 & 1.1525     & 1.3 GB \\
      BrainGB (DTI)  
        & 53.2 & 62.0 & 2.7137     & 1.3 GB
        & 56.9 & 63.1 & 1.0483     & 1.2 GB \\
      IGS (fMRI)  
        & 64.3 & 75.1 & 5.3924     & \underline{0.7 GB}
        & 67.2 & 79.9 & 4.8293     & \underline{0.7 GB} \\
      IGS (DTI)  
        & 61.5 & 74.4 & 4.9221     & 0.7 GB
        & \underline{81.3} & \underline{87.3} & 4.6833     & 0.7 GB \\
    \midrule
      Cross‐GNN         
        & 81.2 & \underline{86.8} & 21.196     & 3.2 GB  
        & 72.7 & 84.7 & 18.961     & 3.1 GB \\
      Mocha‐GCN   
        & 75.9 & 81.0 & 5.8893     & 2.0 GB 
        & 71.3 & 84.2 & 4.2939     & 1.9 GB \\
      RH‐BrainFS  
        & \textbf{85.4} & 66.8 & 24.682     & 5.8 GB
        & 75.0 & 72.7 & 24.144     & 5.8 GB \\
      AL‐NEGAT    
        & 78.1 & 61.3 & 69.053     & 4.6 GB 
        & 77.9 & 60.1 & 68.042     & 4.6 GB \\
    \midrule
      \textbf{BrainMAP}
        & \underline{82.3} & \textbf{91.5} 
          & \underline{0.0243} & 1.5 GB  
        & \textbf{86.2} & \textbf{93.8} 
          & \underline{0.0023} & 1.4 GB \\
      \bottomrule
    \end{tabular}
  \end{adjustbox}
\end{table}

\section{Experiments}
\subsection{Datasets and Data Preprocessing}
We conducted experiments on two public datasets: ADNI \cite{petersen2010alzheimer} and PPMI \cite{marek2011parkinson}, each providing functional (fMRI) and structural (DTI) data. ADNI included 407 participants (190 normal controls (NC), 170 with mild cognitive impairment (MCI), 47 diagnosed with AD). PPMI comprised 158 participants (40 NC, 69 prodromal phase individuals, 49 diagnosed with PD). Each participant provided one fMRI and one DTI scan, covering 90 ROIs. Functional connectivity was extracted using GRETNA \cite{10.3389/fnhum.2015.00386} and the AAL atlas \cite{tzourio2002automated} with Pearson correlation analysis on regional time-series signals. Structural connectivity was derived using PANDA \cite{cui2013panda} for preprocessing and fractional anisotropy metrics following segmentation by the AAL atlas template.
\begin{figure}[t]
  \centering
  \includegraphics[width=\columnwidth]{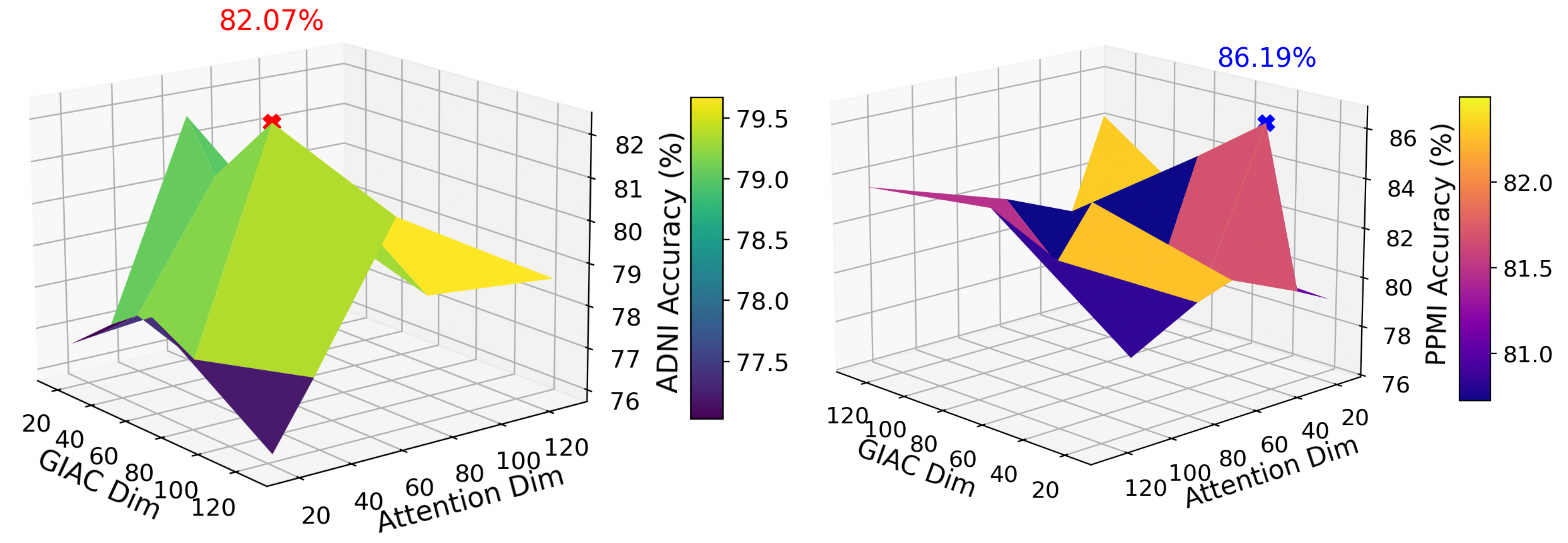}
  \caption{BrainMAP classification accuracy with CNA-GIAC hyperparameters on (1) ADNI (left) and (2) PPMI (right)}
  \label{fig:fusion}
\end{figure}
\begin{figure}[t]
  \centering
  \includegraphics[width=0.9\columnwidth]{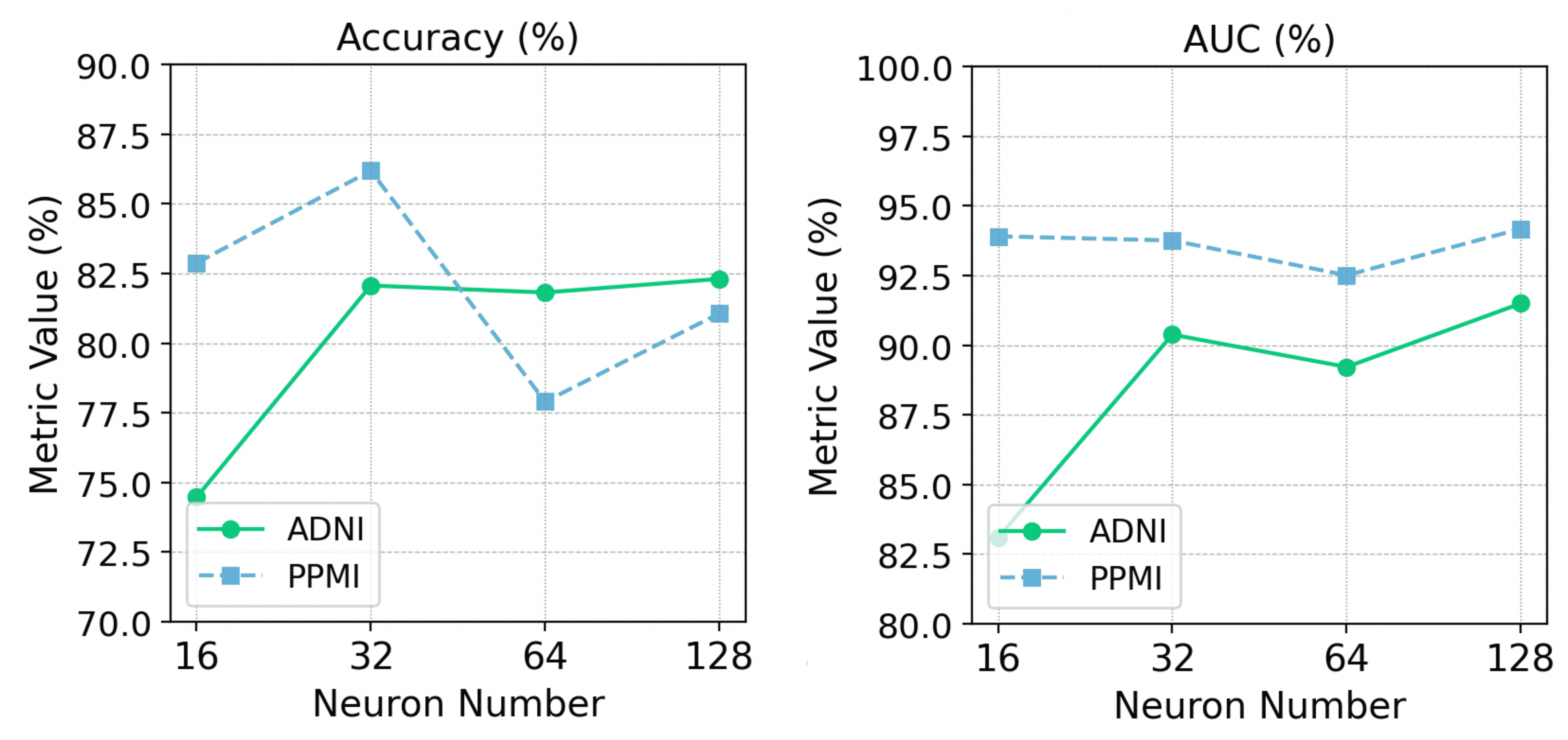}
  \caption{Accuracy and AUC values of BrainMAP under different neuron numbers over two datasets.}
  \label{fig:neuron}
\end{figure}
\subsection{Experimental Settings} 
Predictive performance (ACC, AUC) and efficiency (epoch runtime (seconds), peak memory) are evaluated using five-fold cross-validation, with fold averages reported. All experiments are run on the same GPU setting for a fair efficiency comparison. For ADNI, we use CNA $=64$, GIAC $=32$; for PPMI, both equal $128$. We set masking rate $=0.2$ and feature-group removal $\underline{K}=3$. The graph backbone has $L$ layers of pointwise linear transforms (widths: 128 for ADNI, 32 for PPMI), Laplacian positional encodings, and a GCN layer with dropout 0.1. Models are trained with Adam (learning rate $=0.003$, weight decay $=0.0005$) in batches of 32.


\subsection{Model Comparison}
\subsubsection{Baselines} 
To evaluate BrainMAP’s performance, we benchmark it against diverse graph learning models. Classical methods include GCN and GAT, fundamental frameworks for graph learning. Unimodal models include BrainGB \cite{cui2022braingb}, a unified brain graph benchmark; BrainGNN \cite{li2021braingnn}, which leverages ROI-aware graph convolution; and IGS \cite{li2023interpretable}, which sparsifies graphs to enhance classification. Multimodal baselines include RH-BrainFS \cite{ye2023rh}, using random hypergraph sampling for hierarchical feature selection; AL-NEGAT \cite{chen2022adversarial}, generating adversarial brain-graph negatives; Cross-GNN \cite{yang2023mapping}, embedding structural and functional connectomes into a shared latent space; and MochaGCN \cite{xie2024multimodal}, which fuses DTI and fMRI through Poincaré Fréchet mean convolutions and contrastive fusion in hyperbolic space.

\begin{table}[ht]
  \centering
  \caption{Ablation study performance on the ADNI dataset when omitting AGSF and AGIF}
  \label{tab:ablation_adni}
  \scriptsize
  \setlength{\tabcolsep}{5pt}
  \begin{adjustbox}{width=\columnwidth,center}
    \begin{tabular}{c|cccc}
      \toprule
      \textbf{Method}
        & \textbf{ACC}        
        & \textbf{AUC}          
        & \textbf{Time (s)}                
        & \textbf{Memory} \\ 
      \midrule
      w/o AGSF 
        & \textbf{82.6 ± 3.1} 
        & \underline{88.8 ± 5.0}  
        & 0.0854     
        & 3.1 GB  \\

      w/o AGIF
        & \underline{72.5 ± 5.0} 
        & 84.0 ± 5.8  
        & \textbf{0.0017}    
        & \textbf{0.7 GB}  \\

      \midrule
      \textbf{BrainMAP} 
        & \underline{82.3 ± 3.5} 
        & \textbf{91.5 ± 3.9}
        & \underline{0.0243}     
        & \underline{1.5 GB} \\
      \bottomrule
    \end{tabular}
  \end{adjustbox}
\end{table}

\subsubsection{Result Comparison}
Table \ref{tab:compare} summarizes our results on the ADNI and PPMI cohorts. On ADNI, BrainMAP achieves 82.3\% accuracy and 91.5\% AUC in 0.0243 seconds per epoch and 1.5 GB of GPU memory. Although a simple GAT model trains marginally faster, and GCN, GAT, BrainGB and IGS use slightly less memory, BrainMAP runs nearly as quickly as the fastest unimodal methods, despite using roughly 700 MB more memory, and yet exceeds their accuracy and AUC. Among all multimodal baselines, BrainMAP demonstrates the best balance of time and space efficiency. Against the performance leading multimodal baseline RH-BrainFS, BrainMAP concedes about 3\% accuracy but exceeds its AUC, demonstrating better overall efficiency. On PPMI, BrainMAP sets a new benchmark in performance metrics, all while maintaining minimal runtime and memory use. Overall, classical models are fast and lean but underperform; unimodal models improve accuracy with moderate resources, and multimodal methods further boost it at high cost. By leveraging multimodal data and disease subnetworks, BrainMAP outperforms classical and unimodal models, matching leading multimodal methods with less computation.


\subsection{Parameter Study}
To ensure our model’s robustness, we conducted 2 separate parameter studies including: the CNA-GIAC hidden dimension, and the number of neurons in the GCN classifier.

\paragraph{CNA-GIAC Dimensionality Interaction}
Demonstrates in Fig. \ref{fig:fusion}, we measured how CNA and GIAC dimensions jointly influence performance. On ADNI, accuracy peaks at 82.07\% when both dimensions are 64; on PPMI, it peaks at 86.19\% at 32 for both. ADNI’s accuracy varies by about 6\% across the grid, whereas PPMI spans nearly 10\%, indicating ADNI is more tolerant of fusion-size changes. These findings highlight the need for moderate embedding sizes to balance representational richness and generalization.
\paragraph{GCN Neuron Number Effect}
We examine how varying the number of neurons in the hidden layer of the classifier affects BrainMAP’s performance, as illustrated in Fig. \ref{fig:neuron}. BrainMAP’s performance remains consistently high once a moderate size is used. Although smaller widths can limit capacity, all configurations with 32 or more neurons deliver similar performance, demonstrating that the model is both stable and robust to variations in hidden‐layer width.


\subsection{Ablation Study} In this section, we isolate the contributions of (1) AGSF and (2) AGIF to study about the effectiveness of our proposed technique. The results are reported in Table \ref{tab:ablation_adni}.

\paragraph{The effectiveness of AGSF} To assess the impact of AGSF module, we compared BrainMAP with and without subgraph filtering (i.e. falls back to ${R_0}$ ). The results reveal that in ADNI observed only a slight increase of 0.3\% in accuracy. From an efficiency standpoint, AGSF slashes memory use by over 45\% and cuts per‐epoch runtime by more than 25\% on ADNI, dramatically reducing resource demands without sacrificing generalization. Overall, this confirms our proposal that pruning out irrelevant regions imposes minimal cost to predictive metrics while preserving the model’s ability to distinguish subtle disease patterns.

\paragraph{The effectiveness of AGIF}  
To assess GIF’s contribution, we compared BrainMAP with a simple element-wise product of linearly projected features. This simpler fusion reduced memory use by 45\% and sped up training by 40\%, but caused a significant performance drop. In contrast, GIF preserves near-optimal separation with a good balance of complexity and performance, highlighting its importance for modeling complex intermodal interactions in neurodegenerative disease detection.

\subsection{Model Explainability Analysis}

\begin{figure}[t]
  \centering
  \includegraphics[width=0.9\columnwidth]{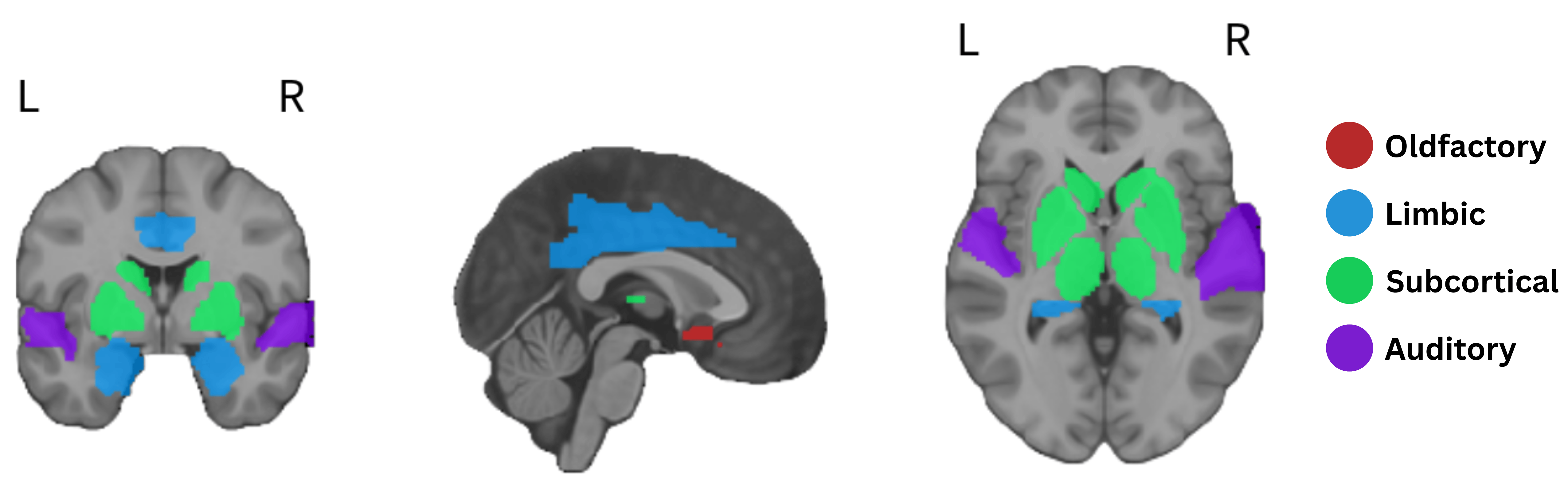}
  \caption{Disease-relevant regions from ADNI fMRI data}
  \label{fig:adni_fmri}
\end{figure}

\begin{figure}[t]
  \centering
  \includegraphics[width=0.9\columnwidth]{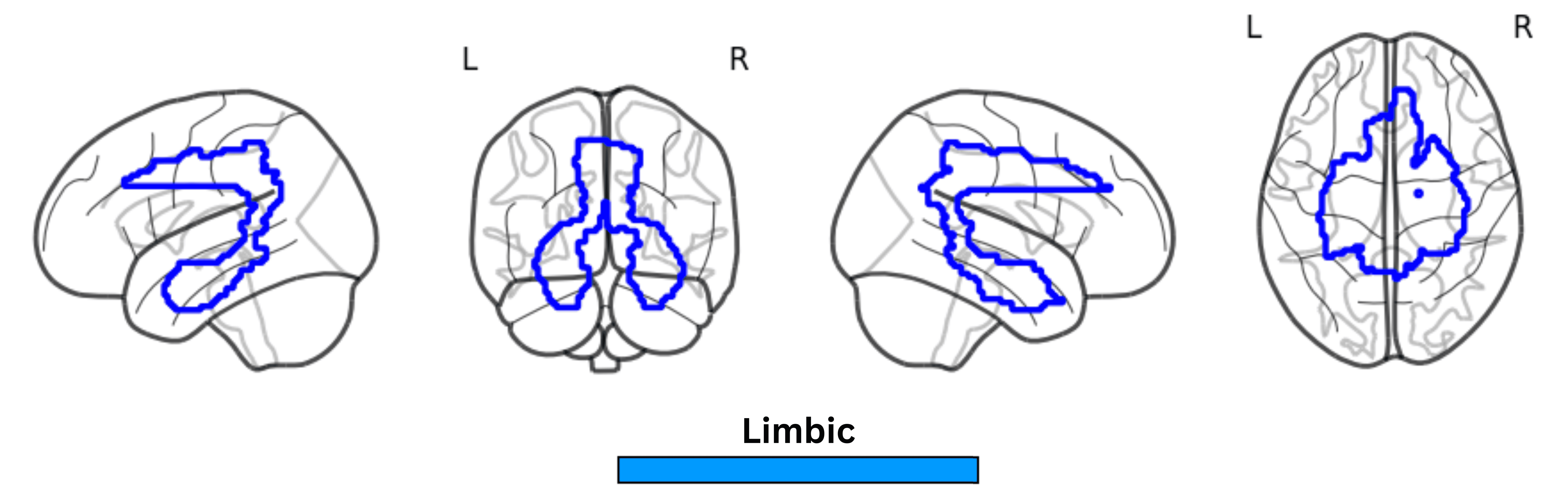}
  \caption{Disease-relevant regions from ADNI DTI data}
  \label{fig:adni_dti}
\end{figure}

Taking ADNI as an illustrative example (Fig. \ref{fig:adni_fmri}, Fig. \ref{fig:adni_dti}), AGSF highlights subcortical hubs, limbic structures, and the olfactory and auditory pathways in fMRI and emphasizes limbic in DTI data, which are the regions that consistently altered early in Alzheimer’s. In particular, the subcortical and limbic areas glow as key foci, alongside weakened olfactory tracts and auditory circuits, mirroring clinical findings of prodromal Alzheimer’s and mild cognitive impairment \cite{forno2023thalamic,seoane2024subcortical}. By focusing on these biologically meaningful circuits, our model not only achieves strong predictive power but also offers transparent, clinically relevant insights into Alzheimer’s progression.


\section{Conclusion and Future Work}
This paper presents BrainMAP as a state-of-the-art multimodal framework that combines computational efficiency with strong predictive accuracy, enabling practical clinical use. In future work, we plan to incorporate streaming temporal graph neural networks for real-time monitoring of brain network dynamics to support clinical decision-making.


\bibliographystyle{IEEEtran}

\addtolength{\textheight}{-12cm}   





\bibliography{reference}

\begin{thebibliography}{10}
\providecommand{\url}[1]{#1}
\csname url@samestyle\endcsname
\providecommand{\newblock}{\relax}
\providecommand{\bibinfo}[2]{#2}
\providecommand{\BIBentrySTDinterwordspacing}{\spaceskip=0pt\relax}
\providecommand{\BIBentryALTinterwordstretchfactor}{4}
\providecommand{\BIBentryALTinterwordspacing}{\spaceskip=\fontdimen2\font plus
\BIBentryALTinterwordstretchfactor\fontdimen3\font minus \fontdimen4\font\relax}
\providecommand{\BIBforeignlanguage}[2]{{%
\expandafter\ifx\csname l@#1\endcsname\relax
\typeout{** WARNING: IEEEtran.bst: No hyphenation pattern has been}%
\typeout{** loaded for the language `#1'. Using the pattern for}%
\typeout{** the default language instead.}%
\else
\language=\csname l@#1\endcsname
\fi
#2}}
\providecommand{\BIBdecl}{\relax}
\BIBdecl

\bibitem{cui2022braingb}
H.~Cui, W.~Dai, Y.~Zhu, X.~Kan, A.~A.~C. Gu, J.~Lukemire, L.~Zhan, L.~He, Y.~Guo, and C.~Yang, ``Braingb: a benchmark for brain network analysis with graph neural networks,'' \emph{IEEE transactions on medical imaging}, vol.~42, no.~2, pp. 493--506, 2022.

\bibitem{ye2023rh}
H.~Ye, Y.~Zheng, Y.~Li, K.~Zhang, Y.~Kong, and Y.~Yuan, ``Rh-brainfs: regional heterogeneous multimodal brain networks fusion strategy,'' \emph{Advances in Neural Information Processing Systems}, vol.~36, pp. 59\,286--59\,303, 2023.

\bibitem{yang2023mapping}
Y.~Yang, C.~Ye, X.~Guo, T.~Wu, Y.~Xiang, and T.~Ma, ``Mapping multi-modal brain connectome for brain disorder diagnosis via cross-modal mutual learning,'' \emph{IEEE Transactions on Medical Imaging}, vol.~43, no.~1, pp. 108--121, 2023.

\bibitem{xie2024multimodal}
C.~Xie, W.~Zhou, C.~Peng, A.~N. Hoshyar, C.~Xu, U.~Naseem, and F.~Xia, ``Multimodal hyperbolic graph learning for alzheimer’s disease detection,'' in \emph{Australasian Joint Conference on Artificial Intelligence}.\hskip 1em plus 0.5em minus 0.4em\relax Springer, 2024, pp. 390--403.

\bibitem{song2022multicenter}
X.~Song, F.~Zhou, A.~F. Frangi, J.~Cao, X.~Xiao, Y.~Lei, T.~Wang, and B.~Lei, ``Multicenter and multichannel pooling gcn for early ad diagnosis based on dual-modality fused brain network,'' \emph{IEEE Transactions on Medical Imaging}, vol.~42, no.~2, pp. 354--367, 2022.

\bibitem{moon2024feature}
H.~S. Moon, A.~Mahzarnia, J.~Stout, R.~J. Anderson, Z.~Y. Han, J.~T. Tremblay, C.~T. Badea, and A.~Badea, ``Feature attention graph neural network for estimating brain age and identifying important neural connections in mouse models of genetic risk for alzheimer’s disease,'' \emph{Imaging Neuroscience}, vol.~2, pp. 1--22, 2024.

\bibitem{li2023interpretable}
G.~Li, M.~Duda, X.~Zhang, D.~Koutra, and Y.~Yan, ``Interpretable sparsification of brain graphs: Better practices and effective designs for graph neural networks,'' in \emph{Proceedings of the 29th ACM SIGKDD Conference on Knowledge Discovery and Data Mining}, 2023, pp. 1223--1234.

\bibitem{chen2022adversarial}
Y.~Chen, J.~Yan, M.~Jiang, T.~Zhang, Z.~Zhao, W.~Zhao, J.~Zheng, D.~Yao, R.~Zhang, K.~M. Kendrick \emph{et~al.}, ``Adversarial learning based node-edge graph attention networks for autism spectrum disorder identification,'' \emph{IEEE Transactions on Neural Networks and Learning Systems}, 2022.

\bibitem{li2021braingnn}
X.~Li, Y.~Zhou, N.~Dvornek, M.~Zhang, S.~Gao, J.~Zhuang, D.~Scheinost, L.~H. Staib, P.~Ventola, and J.~S. Duncan, ``Braingnn: Interpretable brain graph neural network for fmri analysis,'' \emph{Medical Image Analysis}, vol.~74, p. 102233, 2021.

\bibitem{tzourio2002automated}
N.~Tzourio-Mazoyer, B.~Landeau, D.~Papathanassiou, F.~Crivello, O.~Etard, N.~Delcroix, B.~Mazoyer, and M.~Joliot, ``Automated anatomical labeling of activations in spm using a macroscopic anatomical parcellation of the mni mri single-subject brain,'' \emph{Neuroimage}, vol.~15, no.~1, pp. 273--289, 2002.

\bibitem{petersen2010alzheimer}
R.~C. Petersen, P.~S. Aisen, L.~A. Beckett, M.~C. Donohue, A.~C. Gamst, D.~J. Harvey, C.~Jack~Jr, W.~J. Jagust, L.~M. Shaw, A.~W. Toga \emph{et~al.}, ``Alzheimer's disease neuroimaging initiative (adni) clinical characterization,'' \emph{Neurology}, vol.~74, no.~3, pp. 201--209, 2010.

\bibitem{marek2011parkinson}
K.~Marek, D.~Jennings, S.~Lasch, A.~Siderowf, C.~Tanner, T.~Simuni, C.~Coffey, K.~Kieburtz, E.~Flagg, S.~Chowdhury \emph{et~al.}, ``The parkinson progression marker initiative (ppmi),'' \emph{Progress in neurobiology}, vol.~95, no.~4, pp. 629--635, 2011.

\bibitem{park2023deep}
C.~Park, W.~Jung, and H.-I. Suk, ``Deep joint learning of pathological region localization and alzheimer’s disease diagnosis,'' \emph{Scientific reports}, vol.~13, no.~1, p. 11664, 2023.

\bibitem{ye2023explainable}
Z.~Ye, Y.~Qu, Z.~Liang, M.~Wang, and Q.~Liu, ``Explainable fmri-based brain decoding via spatial temporal-pyramid graph convolutional network,'' \emph{Human Brain Mapping}, vol.~44, no.~7, pp. 2921--2935, 2023.

\bibitem{zheng2024bpi}
K.~Zheng, S.~Yu, L.~Chen, L.~Dang, and B.~Chen, ``Bpi-gnn: Interpretable brain network-based psychiatric diagnosis and subtyping,'' \emph{NeuroImage}, vol. 292, p. 120594, 2024.

\bibitem{xing2024adaptive}
T.~Xing, Y.~Dou, X.~Chen, J.~Zhou, X.~Xie, and S.~Peng, ``An adaptive multi-graph neural network with multimodal feature fusion learning for mdd detection,'' \emph{Scientific Reports}, vol.~14, no.~1, p. 28400, 2024.

\bibitem{wei20254d}
Y.~Wei, Y.~Zhang, X.~Xiao, T.~Wang, X.~Wang, and V.~D. Calhoun, ``4d multimodal co-attention fusion network with latent contrastive alignment for alzheimer's diagnosis,'' \emph{arXiv preprint arXiv:2504.16798}, 2025.

\bibitem{qu2025integrated}
G.~Qu, Z.~Zhou, V.~D. Calhoun, A.~Zhang, and Y.-P. Wang, ``Integrated brain connectivity analysis with fmri, dti, and smri powered by interpretable graph neural networks,'' \emph{Medical Image Analysis}, p. 103570, 2025.

\bibitem{peng2024adaptive}
C.~Peng, M.~Liu, C.~Meng, S.~Yu, and F.~Xia, ``Adaptive brain network augmentation based on group-aware graph learning,'' in \emph{The Second Tiny Papers Track at ICLR 2024}, 2024.

\bibitem{10.3389/fnhum.2015.00386}
J.~Wang, X.~Wang, M.~Xia, X.~Liao, A.~Evans, and Y.~He, ``Gretna: a graph theoretical network analysis toolbox for imaging connectomics,'' \emph{Frontiers in Human Neuroscience}, vol. Volume 9 - 2015, 2015.

\bibitem{cui2013panda}
Z.~Cui, S.~Zhong, P.~Xu, Y.~He, and G.~Gong, ``Panda: a pipeline toolbox for analyzing brain diffusion images,'' \emph{Frontiers in human neuroscience}, vol.~7, p.~42, 2013.

\bibitem{forno2023thalamic}
G.~Forno, M.~Saranathan, J.~Contador, N.~Guillen, N.~Falg{\`a}s, A.~Tort-Merino, M.~Balasa, R.~Sanchez-Valle, M.~Hornberger, and A.~Llad{\'o}, ``Thalamic nuclei changes in early and late onset alzheimer's disease,'' \emph{Current Research in Neurobiology}, vol.~4, p. 100084, 2023.

\bibitem{seoane2024subcortical}
S.~Seoane, M.~van~den Heuvel, {\'A}.~Acebes, and N.~Janssen, ``The subcortical default mode network and alzheimer’s disease: a systematic review and meta-analysis,'' \emph{Brain Communications}, vol.~6, no.~2, p. fcae128, 2024.

\end{thebibliography}

\end{document}